\pdfoutput=1
\documentclass[11pt]{article}

\usepackage{acl}

\usepackage{times}
\usepackage{latexsym}

\usepackage[T1]{fontenc}
\usepackage[utf8]{inputenc}

\usepackage{microtype}

\title{The BEA 2023 Shared Task on Generating AI Teacher Responses in Educational Dialogues}

\author{Anaïs Tack \\
  KU Leuven, imec \\
  \small\texttt{anais.tack@kuleuven.be} \\\And
  Ekaterina Kochmar \\
  MBZUAI \\
  \small\texttt{ekaterina.kochmar@mbzuai.ac.ae} \\\AND
  Zheng Yuan \\
  King’s College London \\
  \small\texttt{zheng.yuan@kcl.ac.uk} \\\And
  Serge Bibauw \\
  Universidad Central del Ecuador \\
  \small\texttt{sbibauw@uce.edu.ec} \\\And
  Chris Piech\\
  Stanford University\\
  \small\texttt{cpiech@stanford.edu}\\}

\usepackage{amsmath}
\usepackage{cleveref}

\usepackage{tabularx,multirow,arydshln,bigdelim,siunitx}
\usepackage{float} % H option

\usepackage{algorithm}
\usepackage{algpseudocode}
\usepackage{mathtools}

\usepackage{natbib}

\begin{document}
\maketitle
\begin{abstract}

This paper describes the results of the first shared task on the generation of teacher responses in educational dialogues. The goal of the task was to benchmark the ability of generative language models to act as AI teachers, replying to a student in a teacher-student dialogue. Eight teams participated in the competition hosted on CodaLab. They experimented with a wide variety of state-of-the-art models, including Alpaca, Bloom, DialoGPT, DistilGPT-2, Flan-T5, GPT-2, GPT-3, GPT-4, LLaMA, OPT-2.7B, and T5-base. Their submissions were automatically scored using BERTScore and DialogRPT metrics, and the top three among them were further manually evaluated in terms of pedagogical ability based on \citet{tack_ai_2022}. The NAISTeacher system, which ranked first in both automated and human evaluation, generated responses with GPT-3.5 using an ensemble of prompts and a DialogRPT-based ranking of responses for given dialogue contexts. 
Despite the promising achievements of the participating teams, the results also highlight the need for evaluation metrics better suited to educational contexts.

%This paper describes the results of the first shared task on the generation of teacher responses in educational dialogues. The goal of the task was to benchmark the ability of generative language models to act as AI teachers, replying to a student in a teacher-student dialogue. Eight teams participated in the competition hosted on CodaLab. Most teams experimented with a variety of state-of-the-art models, including Alpaca, Bloom, DialoGPT, DistilGPT-2, Flan-T5, GPT-2, GPT-3, GPT-4, LLaMA, OPT-2.7B, and T5-base. Their submissions were automatically scored and ranked using two metrics: BERTScore and DialogRPT. The top three submissions were further manually evaluated and ranked in terms of pedagogical ability based on \citet{tack_ai_2022}. The NAISTeacher system achieved the highest rank on both automated and human evaluation. The system generated responses with GPT-3.5 Turbo using a varied ensemble of prompts, ranked the set of responses with DialogRPT, and produced the best-ranked response for a given dialogue context. Despite the promising results achieved by the participating teams, the competition also highlighted issues with the use of automated evaluation metrics.
\end{abstract}

% [!] put this here so that it will appear on page 2 (where data is described)
\begin{figure*}[h!]
\begin{tabularx}{\linewidth}{|lX|l}
\multicolumn{1}{l}{\textsc{speaker}} & \multicolumn{1}{X}{\textsc{utterance}} & \multicolumn{1}{l}{} \\
\cline{1-2}
\textbf{Teacher:} & Yes, good! And to charge it up, you need to \_\_ it \_\_\_ & \rdelim]{6}{*}[\textsc{dialogue context}]\\
\textbf{Student:} & … &\\
\textbf{Teacher:} & connect to the source of electricity &\\ 
\textbf{Student:} & i understand &\\
\textbf{Teacher:} & plug it \_\_? &\\
\textbf{Student:} & in & \\
\cdashline{1-2}
\textbf{Teacher:} & yes, good. And when the battery is full, you need to \_\_\_\_ (disconnect it) & = \textsc{reference response}\\
\cline{1-2}
\end{tabularx}
\caption{An example of a sample taken from the \textit{Teacher-Student Chatroom Corpus}}
\label{fig:data-sample}
\end{figure*}

\section{Introduction}

Conversational AI offers promising opportunities for education. Chatbots can fulfill various roles—from intelligent tutors to service-oriented assistants—and pursue different objectives, such as improving student skills and increasing instructional efficiency~\citep{wollny_are_2021}. One of the most important roles of an educational chatbot is that of an AI teacher, helping a student improve their skills and providing more opportunities to practice. Recent studies suggest that chatbots have a significant effect on skill improvement, for example, in language learning~\citep{bibauw_dialogue_2022}. Moreover, the advances in Large Language Models (LLMs) open up new opportunities as such models have the potential to revolutionize education and significantly transform the learning and teaching experience.

%Conversational AI offers promising opportunities for education. Chatbots can fulfill various roles (e.g., intelligent tutors and service-oriented assistants) and pursue different objectives (e.g., improving student skills and increasing instructional efficiency) \citep{wollny_are_2021}. Among the different vocations of an educational chatbot, the most prevalent is the AI teacher who helps a student improve their skills and provides more opportunities to practice. Some recent meta-analyses have even reported a significant effect of chatbots on skill improvement, for example, in language learning \citep{bibauw_dialogue_2022}. In addition, current advances in AI and natural language processing have led to the development of conversational agents based on more powerful language models.

% –> Incorporate the following: (1) from Hicke et al.: ``Conversational agents have the potential to revolutionize the teaching landscape by addressing several challenges and enhancing the overall learning experience for both students and educators (Wollny et al., 2021)." & (2) Development of LLMs has taken the world of education by storm, but what is lacking is a better understanding of whether these models provide educationally useful content.
Despite these promising opportunities, the use of powerful generative models as a foundation for downstream tasks presents several crucial challenges, in particular when such tasks may have real social impact. Specifically, in the educational domain, it is important to determine how solid that foundation is. \citet[67-72]{bommasani_opportunities_2021} stress that if we want to put such models into practice as AI teachers, it is crucial to determine whether they can (a) speak to students like a teacher, (b) understand students, and (c) help students improve their understanding. Following these desiderata,~\citet{tack_ai_2022} formulated the AI Teacher Test Challenge: {\em How can we test whether state-of-the-art generative models are good AI teachers, capable of replying to a student in an educational dialogue?}

%Despite these promising opportunities, the use of powerful generative models as a foundation for downstream tasks presents several crucial challenges. In the educational domain, in particular, it is important to determine whether that foundation is solid or flimsy.~\citet{bommasani_opportunities_2021} (pp. 67-72) stressed that if we want to put these models into practice as AI teachers, it is imperative to determine whether they can (a) speak to students like a teacher, (b) understand students, and (c) help students improve their understanding. Following these desiderata,~\citet{tack_ai_2022} formulated the AI teacher test challenge: How can we test whether state-of-the-art generative models are good AI teachers, capable of replying to a student in an educational dialogue?

Building on the AI Teacher Test Challenge, we have organized the first shared task on the generation of teacher language in educational dialogues. The goal of this task is to explore the potential of NLP and AI methods to generate teacher responses in the context of real-world teacher–student interactions. Interaction samples were extracted from the {\em Teacher Student Chatroom Corpus}~\citep{caines_teacherstudent_2020,caines_teacherstudent_2022}, with each training sample consisting of a dialogue context (i.e., several rounds of teacher-student utterances) and the teacher's response. For each test sample, participants were asked to submit their best generated teacher response.

As the purpose of this task was to benchmark the ability of generative models to act as AI teachers responding to a student in a teacher-student dialogue, the submissions were first ranked according to popular BERTScore and DialogRPT metrics. The top three submissions were then selected for further human evaluation. During this manual evaluation, the raters compared a pair of ``teacher" responses along three dimensions: speaking like a teacher, understanding a student, and helping a student~\citep{tack_ai_2022}.

\section{Materials and Methods}

The shared task used data from the \textit{Teacher-Student Chatroom Corpus} (TSCC)~\citep{caines_teacherstudent_2020,caines_teacherstudent_2022}. This corpus comprises data from several chatrooms in which an English as a second language (ESL) teacher interacts with a student to work on a language learning exercise and assess the student’s English language proficiency.

\subsection{Data Samples}
\label{sec:data}

Several samples were taken from each dialogue in the corpus. Each sample consisted of several sequential teacher-student turns (i.e., the preceding dialogue context) and ended with a teacher utterance (i.e., the reference response). Figure \ref{fig:data-sample} shows an example of a sample taken from the corpus. As can be seen from this example, the samples were quite short, counting at most $100$ tokens. The length of each sample had to be capped at this specific limit in order to comply with the copyright license and terms of use of the corpus, even though this restricted context inevitably posed an important limitation on training and testing.

\subsubsection{Extraction}

The samples were extracted with the following method. 
%using \Cref{alg:data-sampling}. 
For each dialogue in the corpus, the sequence of utterances was iterated from the first to the last. If the speaker of an utterance at the current position was a teacher, the utterance was a potential reference response. In that case, a contextual window sequence was created for the reference candidate by recursively backtracking through the dialogue and adding the preceding utterances until the limit of $100$ tokens was reached. Each utterance was tokenized with spaCy's default tokenizer for English.\footnote{\url{https://spacy.io/api/tokenizer}} Once extracted, the sequence was added to the set of samples for the dialogue on the condition that it had at least two utterances and more than one speaker. For example, if the teacher initiated the conversation, the algorithm would extract a window with only one speaker and no preceding utterances. Because this instance would not have been informative, it was ignored and was not added to the set of data samples. A total of $7,047$ data samples were extracted from the original dataset.

\begin{figure}[t!]
\centering
\includegraphics[width=\columnwidth]{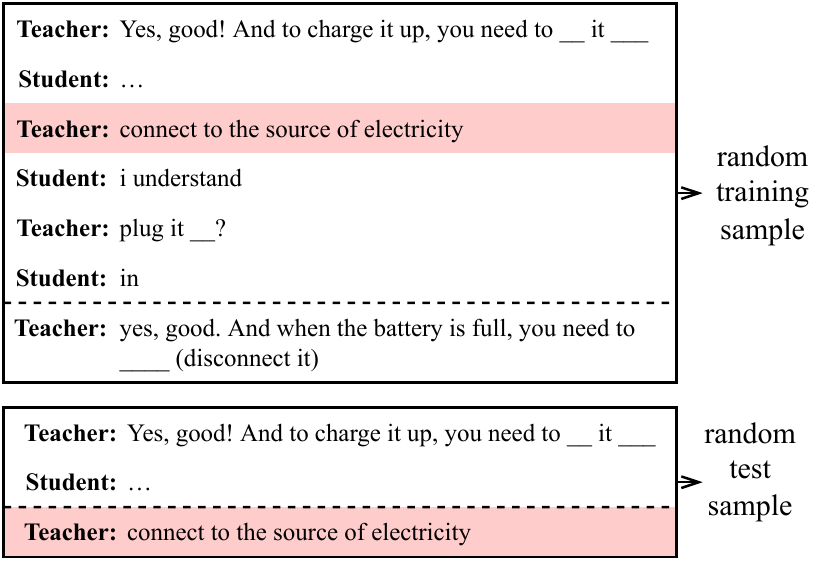}
\caption{An example of a reference in a test sample observed in the context of a training sample}
\label{fig:data-sample-issue-1}
\end{figure}

\subsubsection{Selection}

\begin{figure}[t]
\centering
\includegraphics[width=\columnwidth]{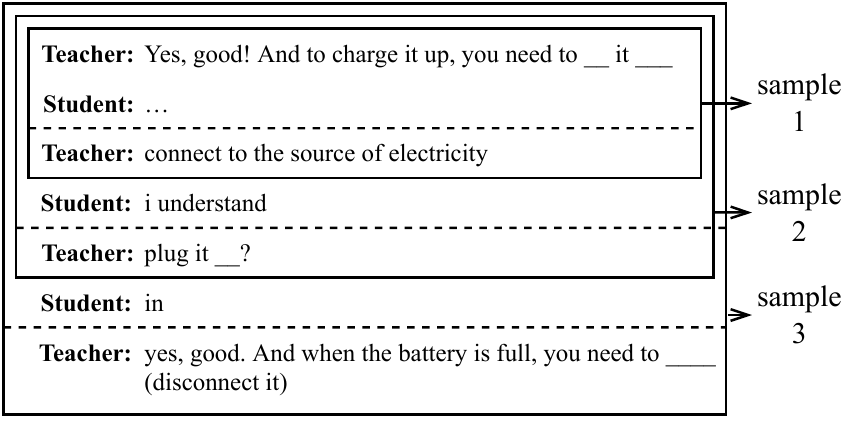}
\caption{An example of a nested or Russian doll-like ensemble of data samples}
\label{fig:data-sample-issue-2}
\end{figure}

Although the extracted data samples could have been randomly divided into training and test samples, such an approach would have been problematic. In fact, it would have been possible for a randomly selected test sample to contain a reference response otherwise observed in the dialog context of {\em another} randomly selected training or test sample (see \Cref{fig:data-sample-issue-1}). A related issue was that the extraction algorithm produced samples that were also part of other samples, resulting in multiple nested or Russian doll-like ensembles (see \Cref{fig:data-sample-issue-2}). Since a test set should never include references seen elsewhere in the data, special attention was paid to data splitting.

The data samples were divided into a training set and a test set with a more complex selection procedure. Three selection criteria were defined: (a) whether the reference response was labeled as \emph{eliciting} and/or \emph{scaffolding} (`yes' $\Rightarrow$ better), (b) the number of distinct types of conversational organization (e.g., opening, closing, eliciting, scaffolding, and revision) that were added as labels to the reference response (more $\Rightarrow$ better), and (c) the total number of tokens in the sample (more $\Rightarrow$ better). The extracted data samples contained $1,400$ nested ensembles (cf. \Cref{fig:data-sample-issue-2}). The samples in each ensemble were sorted based on the three above criteria, and for each ensemble, only the best sample was selected. The remaining $4,864$ samples were assigned to $2,457$ training and $273$ test slots with the \href{https://docs.scipy.org/doc/scipy/reference/generated/scipy.optimize.linear_sum_assignment.html}{Hungarian algorithm} \cite{kuhn_hungarian_1955} based on the above criteria. Once the assignment was done, the training and test sets were verified for potential conflicts (cf. \Cref{fig:data-sample-issue-1}). Conflicts were resolved using the above criteria to choose the best sample among conflicting samples. Then, the assignment was run again on the remaining samples until no more conflicts could be detected. After the assignment was completed, the nested data samples that were discarded before were used to increase the size of the training set, provided that they were not in conflict with the test set. Finally, the training set was randomly split into a 90\% training and 10\% held-out set. The number of samples included in the training and test sets is shown in \Cref{tab:data-statistics}.

\begin{table}[ht]
\centering
\begin{tabular}{|l|l|r|}
\hline
\multirow{3}{*}{Training set} & \multicolumn{2}{l|}{$3,052$} \\
\cline{2-3}
     & $90\%$ training & $2,747$ \\
     & $10\%$ held-out & $305$ \\
\hline
Test set & \multicolumn{2}{l|}{$273$} \\
\hline
\end{tabular}
\caption{Number of training and test samples}
\label{tab:data-statistics}
\end{table}

\subsection{Competition}

The shared task was hosted as an \href{https://codalab.lisn.upsaclay.fr/competitions/11705}{online competition} on the CodaLab platform \citep{codalab_competitions}. Anyone participating in the shared task filled out a registration form, signed to comply with the terms and conditions of the shared task and the licensed TSCC data, and registered on the CodaLab platform. Participants could only be part of one team, while a team could have one or more members.

\subsubsection{Phases}

The competition was run in two phases: a development phase and an evaluation phase. All deadlines were set to 23:59 Anywhere on Earth (UTC-12). Since CodaLab uses Coordinated Universal Time, all deadlines on the platform were adapted accordingly (i.e.,~set to the next day at 11:59 am UTC). % Wondering if these two sentences about AOE should maybe better be removed (Serge)

The development phase started on March 24, 2023, and ended on April 30, 2023. At the start of the development phase, participants received the training and held-out development data, which were available on the CodaLab platform. During the development phase, participants could submit their results for the held-out data and view their scores on the anonymized leaderboard. %\ParticipantsRegisteredCodaLabPhaseOne{} 
Sixty-three people completed the registration form and registered on the CodaLab platform. Among them, $12$ people actively participated in the development phase and submitted results on the held-out data. Three people submitted to the development phase after the evaluation phase had already started. In the end, $10$ participants made at least one successful submission to the development phase. In total, $17$ successful submissions were received ($M_\text{submissions}=1.7$ per participant). The leaderboard featured only the best successful submission per participant (see the metrics described in \Cref{sec:evaluation-metrics}).

The evaluation phase started on May 1st, 2023, and ended on May 5th, 2023. At the start of the evaluation phase, participants received the test data, which were available on the CodaLab platform. During the evaluation phase, participants could submit their results on the test data and view their scores on the anonymized leaderboard. Furthermore, six people completed the registration form and registered on the CodaLab platform. Nineteen people actively participated in the evaluation phase and submitted their results on the test data. In the end, 10 participants from eight teams made at least one successful submission to the evaluation phase. In total, 19 successful submissions were received ($M_\text{submissions}=1.9$ per participant). Again, the leaderboard featured only the best successful submission per participant (see the metrics described in \Cref{sec:evaluation-metrics}).

It should be noted that some people showed interest in the shared task but did not fully participate. Fifteen people filled in the registration form but did not request to join on the platform before the deadline, whereas %\ParticipantsCodaLabNotRegistered{} 
$18$ people requested to join on CodaLab but did not fill out the registration form. As a result, they could not be accepted into the competition because they did not sign to comply with the terms and conditions.

\subsubsection{Teams and Systems}

% ekaterina zheng
% ekaterina: Done

% Introduction/Summary: how many teams, what are the most popular approaches
Eight teams made at least one successful submission to the final evaluation phase. The approaches taken by the teams were based on a range of state-of-the-art large language models (LLMs), including Alpaca (Team RETUYT-InCo),
Bloom (RETUYT-InCo), 
DialoGPT (Cornell), 
DistilGPT-2 (DT), 
Flan-T5 (teams Cornell and TanTanLabs), 
GPT-2 (Cornell and Data Science-NLP-HSG),
GPT-3 (NBU), 
GPT-3.5 Turbo (NAIST and aiitis),
GPT-4 (Cornell),
LLaMA (RETUYT-InCo),
OPT-2.7B (RETUYT-InCo), and 
T5-base (Data Science-NLP-HSG). 
In addition, all teams experimented with zero- and few-shot learning, fine-tuning, and various prompting strategies. Several teams applied reinforcement learning (RL) (Cornell and Data Science-NLP-HSG), and some developed customized approaches to post-processing (NAIST) and data-driven prompt engineering (aiitis).
All these approaches are summarized below and further detailed in the corresponding system papers.

% --> 1st @anais
\paragraph{Team NAIST}

\citet{vasselli_naisteacher_2023} participated in the shared task with the {\sc NAISTeacher} system, built on a pre-trained GPT-3.5 Turbo \cite{brown2020language}. They experimented with, on the one hand, zero-shot prompts and, on the other hand, few-shot prompts using either handcrafted, generative, or iterative examples of teacher responses. They also experimented with asking the model to generate either one response or several possible responses and compared the performance of their system in two settings: {\em teacher replies} (i.e., when the generated teacher utterance followed a student utterance) and {\em teacher continuations} (i.e., when the generated teacher utterance followed a teacher utterance). Finally, the candidate responses were post-processed (with a profanity filter and regular expressions) and reranked with DialogRPT (see shared task metrics in \Cref{sec:evaluation-metrics}) to select the best response to be submitted for each test sample.

% --> 2nd
\paragraph{Team NBU}

\citet{adigwe_promptbased_2023} participated in the shared task with the ADAIO system. They evaluated several GPT-3 models~\citep{brown2020language}, designed various zero-shot and few-shot prompts to generate teacher responses, and also fine-tuned the models on the TSCC corpus. Additionally, the team experimented extensively with various aspects of response generation by considering the roles of the participants, the teaching approaches taken by the tutor, and the specific teaching goals. The responses submitted to the competition were generated using a few-shot prompt-based method based on the {\it text-davinci-003} model.

% --> 3rd + 6th
\paragraph{Team Cornell}

\citet{hicke_assessing_2023} experimented with several generative models and various approaches, including few-shot in-context learning with GPT-4, fine-tuning of GPT-2 \citep{radford2019language} and DialoGPT \citep{zhang2019dialogpt}, and fine-tuning of Flan-T5 \citep{chung2022scaling} with RL~\citep{ramamurthy2022reinforcement} to optimize for pedagogical quality. Among these, GPT-4 achieved the best results on the shared task evaluation metrics (see~\Cref{sec:evaluation-metrics}). The team made two submissions to the leaderboard: one submission with responses generated by GPT-4, and another submission that included the same responses with a teacher prefix prepended to each of them (\texttt{"teacher:~<response>"}). To distinguish between these submissions, the latter is referred to as GPT-4\textsuperscript{(TP)} where TP stands for teacher prefix.

% --> 4th
\paragraph{Team aiitis}

\citet{omidvar_empowering_2023} introduced the Semantic In-Context Learning ({\sc s-icl}) model. Their aim was to address the challenges created by the use of out-of-the-box pre-trained LLMs, such as domain adaptivity and the high costs of fine-tuning. Their in-context learning approach consisted of providing an LLM (in this case, ChatGPT with the {\it gpt-3.5-turbo} engine) with a prompt containing an instruction, a few labeled samples, and an unlabeled sample. The {\em semantic} component in the {\sc s-icl} model retrieved sufficiently similar samples from the training set, which were then integrated into the prompt fed to the LLM as labeled samples. The inclusion of relevant conversational samples in the prompt allowed the model to leverage available knowledge to generate teacher responses.

% Paragraph sent by Amin Omidvar, not yet adapted:
%S-ICL, the fourth-ranked method in both development and evaluation phases, uses both semantic search and a large language model (ChatGPT) to generate a response to an inquiry based on previous conversations (available data) without the need for gradient-based training. 
%Their proposed approach first finds the most relevant conversations to the incoming inquiry using semantic search. Then, based on the selected prompt engineering technique, it includes the relevant past conversations as samples inside the prompt and leverages both the power of an LLM and the available knowledge in previous conversations to generate the response.

% --> 5th and 8th

\paragraph{Team RETUYT-InCo}

\citet{baladon_retuytinco_2023} experimented with several open-source LLMs, including LLaMA \citep{touvron2023llama}, Alpaca \citep{taori2023stanford}, OPT-2.7B \citep{gao_dialogue_2020}, and Bloom 3B \citep{scao2022bloom}. They explored fine-tuning techniques by applying the LoRA \citep{hu2021lora} method to the aforementioned LLMs. They tested several prompting strategies, including few-shot and chain-of-thought approaches. Their method consisted of selecting the three most similar conversations from the training data using the $k$-nearest neighbors algorithm. These were then further integrated into the prompt for the few-shot learning scenario. The models submitted to the competition were trained using Alpaca LoRA with the few-shot approach, LLaMA 7B with engineered prompts fine-tuned with LoRA, and fine-tuned OPT-2.7B using preprocessing.

% --> 7th
\paragraph{Team Data Science-NLP-HSG}

\citet{huber_enhancing_2023} presented a simple approach to fine-tuning a language model with RL and used the novel NLPO algorithm~\citep{ramamurthy2022reinforcement} that masks out tokens during inference to direct the model toward generations that maximize a reward function. They used Hugging Face's implementation of the T5-base model~\citep{raffel2020exploring} with $220$ million parameters to generate the responses submitted to the competition.

% --> 9th, Rabin Banjade (U Memphis)
\paragraph{Team DT}

% this is now adapted (Ekaterina)

This team experimented with fine-tuning the DistilGPT-2 model specifically for student–teacher dialogues. They divided the original training data using an $80$/$20$ split and ran a three-epoch training process using the Adam optimizer along with a linear learning rate scheduler on the training subset. The remaining $20\%$ was then used for rigorous evaluation using shared task performance metrics.
% = RESULTS, not methods section
% In these experiments, $P=0.77$, $R=0.74$ and F1-score of $0.75$ were achieved for the BERTScore metrics, and the scores of $0.36$, $0.75$, $0.96$, and $0.29$ were achieved for DialogRPT's updown, human vs. random, human vs. machine, and the final average score, respctively (see Section \ref{sec:evaluation-metrics}).
The team \href{https://huggingface.co/rbnjade1/distilgpt2-finetuned-dialogue}{released their model on Hugging Face} and plans to explore the potential of larger models like GPT-3 and GPT-4 in the educational dialogue domain in the future.\footnote{Written by Rabin Banjade and adapted by the authors.}

%Rabin Banjade participated in the shared task with ...

% Paragraph sent by Rabin Banjade, not adapted:
%For our system, we focused on finetuning the DistilGPT-2 model specifically for student-teacher dialogue. Our training dataset was divided into an 80-20 split, with 80\% allocated for training and the remaining 20\% for validation purposes.  To enhance the model's performance, we employed a three-epoch training process using the Adam optimizer along with a linear learning rate scheduler. Following the training phase, we rigorously evaluated the model using a separate test dataset and computed several performance metrics. The BERTScore precision, recall l, and F1-score achieved notable values of 0.77, 0.74, and 0.75, respectively. Additionally, we utilized the DialogRPT metrics, including updown, human vs. random, human vs. machine, and the final average score. The obtained scores were 0.36, 0.75, 0.96, and 0.29, respectively. Although DistilGPT-2 demonstrates its efficiency by consuming fewer computational resources, we plan to expand our investigation by exploring the potential of larger models like GPT-3 and GPT-4 in this specific domain. Our model is also available in HuggingFace.\footnote{\url{https://huggingface.co/rbnjade1/distilgpt2-finetuned-dialogue}}

% --> 10th, TanTanLabs (Tanay Gahlot)
\paragraph{Team TanTanLabs} 

% this is now adapted (Ekaterina)

This team experimented with a zero-shot approach using Hugging Face's Flan-T5 transformer model, a model instruction-finetuned on a mixture of tasks. 
%The team worked on generating appropriate prompts, decoding techniques for model inference, and parsing the model's output.
Among the many prompting techniques tested, the one that worked best was the prompt used by the authors of the Flan-T5 model: ``Read the dialog and predict the next turn.'' %\n{dialog\_}" 
For model inference, different decoding techniques were tried (greedy, decoding by sampling with temperature, and beam search). Beam search was chosen because it was easy to control. %The sampling techniques generated random outputs even at low temperature.
Customized regular expressions were used to parse the model's output. When the model did not produce any output, the filler word ``Alright'' was used. In the future, the team plans to further experiment with supervised fine-tuning using ``chain of thought'' reasoning instructions.\footnote{Written by Tanay Gahlot and adapted by the authors.}

%Team TanTanLabs~\footnote{This paragraph was written by Tanay Gahlot and adapted by the authors.} participated in the shared task with ...

% One-page description sent by Tanay Gahlot, not yet adapted:
%I used a zero-shot approach with a transformer model (FLAN-T5 on HuggingFace) which was instruction-finetuned on a mixture of tasks. A bulk of work went into:
%- Prompt generator: Generating appropriate prompts for this model
%- Model Inference: Decoding techniques
%- Output parsing: Parsing model’s output and generating teacher's responses.
%1) Prompt generator
%I tried many prompting techniques but the one that worked the best for me was the one which was used by the authors of FLAN while instruction finetuning it: ``Read the dialog and predict the next turn."%\n{dialog\_}"
%2) Model Inference
%I tried the decoding techniques mentioned below:
%- Greedy
%- Decoding by sampling with temperature.
%- Beam search 
%I chose beam search as it was easy to control. The sampling techniques generated random outputs even at low temperature.
%3) Response generation
%In response generation, we parse the model’s output using a regex("teacher$\backslash$s*:$\backslash$s*(.+)<$\backslash$/s>"). When the model doesn’t seem to have any output, we respond with the filler word - ``Alright". 
%Next steps: Supervised fine-tuning using instructions containing  “chain of thought” reasoning.
%Alignment of output using RLHF.

\subsection{Evaluation Procedure}
\label{sec:evaluation-procedure}

The submissions made by the teams described above were evaluated in two stages. During the competition, all submissions were automatically scored with several dialogue evaluation metrics \citep[see][for a comprehensive review]{yeh_comprehensive_2021}. The teams used these metrics to optimize their systems before the end of the competition. After the competition ended, the final submissions were evaluated by human raters. Due to combinatorial constraints imposed by the human evaluation task (see \Cref{sec:human-evaluation}), it was not possible to evaluate manually any number of submissions. For this reason, only the top three submissions ranked by the automated metrics were targeted for human evaluation.

\subsubsection{Evaluation Metrics}
\label{sec:evaluation-metrics}

\citet{yeh_comprehensive_2021} reviewed several dialogue evaluation metrics that operate at the level of individual turns (i.e., generated responses). However, many of these metrics required a complicated installation procedure. The following two metrics were used because they were well known, could be easily installed, and their scores could be reproduced.

\textbf{BERTScore} \citep{zhang*_bertscore_2020} was used as a metric to evaluate each generated response relative to the reference (i.e., teacher) response. The metric matches words in submissions and reference responses by cosine similarity. BERTScore was computed with Hugging Face's \emph{evaluate} package and the \textit{distilbert-base-uncased}\footnote{The hashcode was \textit{distilbert-base-uncased\_L5\_no-idf\_version=0.3.12(hug\_trans=4.28.1)}.} model. The resulting precision, recall, and F1 scores were averaged for all items in the test set.

\textbf{DialogRPT} \citep{gao_dialogue_2020} was used as a reference-free metric to evaluate the generated response with respect to the preceding dialogue context. The metric consists of a set of ranked pre-trained transformer models proposed by Microsoft Research NLP Group. These metrics were aggregated for all items in the test set. The following dialog response ranking models were used:
\begin{description}
\item[updown] likelihood that a response gets the most upvotes (mean of all items)
\item[human vs. rand] likelihood that a response is relevant for the given context (mean of all items)
\item[human vs. machine] likelihood that a response is human-written rather than machine-generated (mean of all test items)
\item[final] weighted ensemble score of all DialogRPT metrics (mean of all items).
%\item[final (best)] weighted ensemble score of all DialogRPT metrics (maximum of all items)
\end{description}

Each submission was ranked from $1$ (highest) to $10$ (lowest) on each individual metric. The overall leaderboard rank was computed as the mean rank on BERTScore F1 and on DialogRPT final average. In the case of a tie, the tiebreaker was the mean rank on the individual scores for BERTScore (precision, recall) and DialogRPT (updown, human vs. rand, human vs. machine).

\subsubsection{Human Evaluation}
\label{sec:human-evaluation}

The top $k=3$ submissions on the leaderboard were further evaluated using pairwise comparative judgments.% maybe remove the footnote below to win some space? (Anais)
\footnote{In pairwise comparative judgments, multiple alternatives are evaluated by systematically assessing them in pairs. Each rater is presented with two alternatives at a time and makes a judgment about which one is better according to some criteria. These judgments are used to compute a relative ranking among the alternatives. This method has already been used for the evaluation of dialogue systems \cite{li_acuteeval_2019} and open-ended natural language generation \cite{pillutla_mauve_2021}.} For each sample in the set of $n=273$ test items, the possible responses were combined in pairs so that the generated responses were either compared with the reference (i.e.,~teacher vs.~AI) or between themselves (i.e.,~AI vs.~AI). This resulted in ${k+1 \choose 2} = 6$ pairs of responses for each test sample. Each pair was assessed by $r=3$ raters, which amounted to a total of $\frac{(k+1)!}{2!(k+1-2)!}r=4,914$ different evaluations. These evaluations were collected via an online Qualtrics survey following a method described in \citet{tack_ai_2022} and further detailed below.

\paragraph{Survey}
In the introductory part of the survey, raters were given a short introduction, a consent form, and an example to familiarize themselves with the task at hand. In the central part of the survey, each rater was presented with a comparative judgment task of 20 items that were randomly and evenly selected from the set of $n$ test samples. Each survey item included a pairwise comparison that was randomly and evenly selected from the ${k+1 \choose 2}$ possible pairs for the chosen test sample. Each survey item had three components: the dialogue context, a comparison of two responses (A or B), and three questions targeting pedagogic abilities (\emph{more likely said by a teacher}, \emph{better understanding the student}, and \emph{helping the student more}). For each question, the rater was asked to choose option A or B. The order of presentation in the pairwise comparison was determined randomly so that any presentation order effects would be avoided.

\paragraph{Raters}
A sample of $298$ raters was recruited from the Prolific crowdsourcing platform. The raters were screened based on several requirements: (a) they were from a majority native English-speaking country,\footnote{Based on the \href{https://assets.publishing.service.gov.uk/government/uploads/system/uploads/attachment_data/file/984675/english-language-v19.0ext.pdf\#page=6}{UK government classification} + Ireland.}
% the following countries were considered: Antigua and Barbuda, Australia, The Bahamas, Barbados, Belize, Canada, Dominica, Grenada, Guyana, Ireland, Jamaica, Malta, New Zealand, St Kitts and Nevis, St Lucia, St Vincent and the Grenadines, Trinidad and Tobago, United Kingdom, United States of America.
(b) their native language was English, and (c) their employment sector was in education and training. The sample of raters was gender-balanced. Five raters were removed because the outlier detection described in \citet{tack_ai_2022} showed that they consistently picked the same option (A or B) for all questions throughout the survey.

\paragraph{Ranking}
For each item in the test set, the possible responses were ranked from $1$ (highest) to $4$ (lowest) for each of the three questions (\emph{more likely said by a teacher}, \emph{understanding the student better}, and \emph{helping the student more}). The rank of each response (i.e., teacher or AI) was estimated with a Bayesian Bradley-Terry model and an HMC-NUTS sampler, as described in \citet{tack_ai_2022}. Based on the set of draws produced by the HMC-NUTS sampler, the mean rank, standard deviation, and $95\%$ highest density intervals (HDI) were computed for each item and for each response.

\section{Results}

\begin{figure}
\centering
\includegraphics[width=\columnwidth]{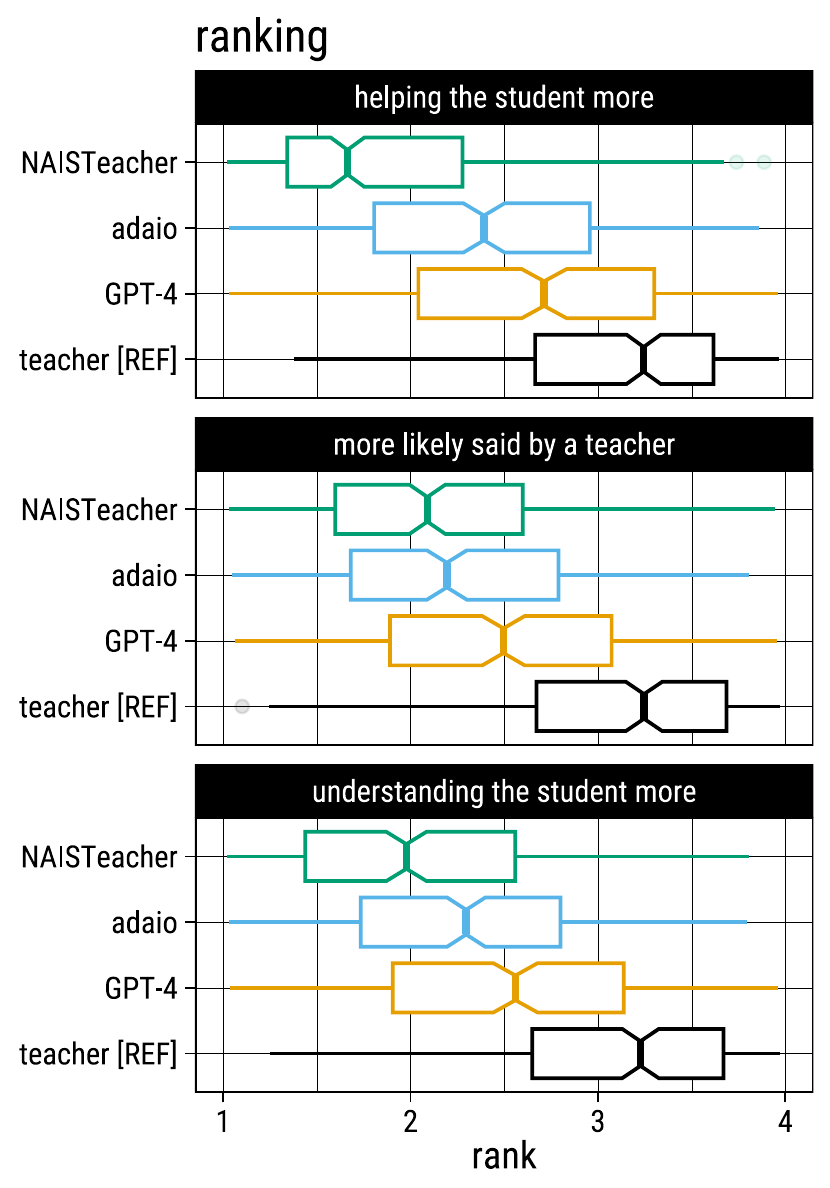}
\caption{Ranking of the top three submissions and the teacher reference after the human evaluation phase.}
\label{fig:human-evaluation}
\end{figure}

The results achieved by the participating teams during the automated evaluation phase are shown in \Cref{tab:automated-metrics}. Those achieved by the top three during the human evaluation phase are shown in \Cref{fig:human-evaluation}. 

\begin{table*}[h]
\begin{tabularx}{\linewidth}{
    |X|>{\raggedright\arraybackslash}X
    |S[table-format=1.2]S[table-format=1.2]||S[table-format=1.2]|
    |S[table-format=1.2]S[table-format=1.2]S[table-format=1.2]||S[table-format=1.2]|c|
}
\hline
\multirow{2}{\hsize}{\textbf{Team}} & \multirow{2}{\hsize}{\textbf{System}}  & \multicolumn{3}{c|}{\textbf{BERTScore}} & \multicolumn{4}{c|}{\textbf{DialogRPT}} & \multirow{2}{*}{\bf Rank} \\
\cline{3-5}\cline{6-9}
{} & {} & {P} & {R} & {\bfseries F1} & {U} & {HvR} & {HvM} & {\bfseries Final} & \\
\hline\hline
\multirow{2}{\hsize}{NAIST} & \multirow{2}{\hsize}{NAISTeacher} & 0.71 & \bfseries 0.71 & \bfseries 0.71 & 0.48 & \bfseries 0.98 & \bfseries 1.00 & 0.46 & \multirow{2}{*}{1.5} \\
{} & {} & {(9)} & {(1)} & {(1)} & {(2)} & {(1)} & {(1)} & {(2)} & \\
\hline
\multirow{2}{\hsize}{NBU} & \multirow{2}{\hsize}{ADAIO} & 0.72 & 0.69 & 0.71 & 0.40 & 0.97 & 0.98 & 0.37 & \multirow{2}{*}{3.0} \\
{} & {} & {(4)} & {(3)} & {(3)} & {(5)} & {(2)} & {(5)} & {(3)} & \\
\hline
\multirow{2}{\hsize}{Cornell} & \multirow{2}{\hsize}{GPT-4\textsuperscript{(TP)}} & 0.71 & 0.69 & 0.70 & \bfseries 0.52 & 0.86 & 0.98  & \bfseries 0.47 & \multirow{2}{*}{3.0} \\
{} & {} & {(7)} & {(2)} & {(5)} & {(1)} & {(8)} & {(2)} & {(1)} & \\
\hline\hline
\multirow{2}{\hsize}{aiitis} & \multirow{2}{\hsize}{S-ICL} & 0.72 & 0.69 & 0.70 & 0.40 & 0.92 & 0.98 & 0.36 & \multirow{2}{*}{4.5}\\
{} & {} & {(3)} & {(5)} & {(4)} & {(4)} & {(5)} & {(4)} & {(5)} &\\
\hline
\multirow{2}{\hsize}{RETUYT-InCo} & \multirow{2}{\hsize}{OPT-2.7B} & \bfseries 0.74 & 0.68 & 0.71 & 0.38 & 0.90 & 0.96 & 0.35 & \multirow{2}{*}{4.5} \\
{} & {} & {(1)} & {(6)} & {(2)} & {(7)} & {(7)} & {(9)} & {(7)} & \\
\hline
\multirow{2}{\hsize}{Cornell} & \multirow{2}{\hsize}{GPT-4} & 0.72 & 0.69 & 0.70 & 0.40 & 0.93 & 0.98 & 0.36 & \multirow{2}{*}{6.0} \\
{} & {} & {(5)} & {(4)} & {(6)} & {(6)} & {(4)} & {(3)} & {(6)} & \\
\hline
\multirow{2}{\hsize}{Data Science-NLP-HSG} & \multirow{2}{\hsize}{Untrained} & 0.72 & 0.63 & 0.67 & 0.41 & 0.93 & 0.95 & 0.37 & \multirow{2}{*}{6.0}\\
{} & {} & {(6)} & {(8)} & {(8)} & {(3)} & {(3)} & {(10)} & {(4)} & \\
\hline
\multirow{2}{\hsize}{RETUYT-InCo} & \multirow{2}{\hsize}{Alpaca} & 0.72 & 0.68 & 0.70 & 0.37 & 0.91 & 0.96 & 0.34 & \multirow{2}{*}{7.5} \\
{} & {} & {(2)} & {(7)} & {(7)} & {(8)} & {(6)} & {(7)} & {(8)} & \\
\hline
\multirow{2}{\hsize}{DT} & \multirow{2}{\hsize}{DistilGPT2}  & 0.67 & 0.62 & 0.64 & 0.36 & {0.75} & 0.96 & 0.29 & \multirow{2}{*}{9.5} \\ 
{} & {} & {(10)} & {(9)} & {(10)} & {(9)} & {(10)} & {(6)} & {(9)} & \\
\hline
\multirow{2}{\hsize}{TanTanLabs} & \multirow{2}{\hsize}{zero-shot-with-filler} & 0.71 & 0.60 & 0.65 & 0.32 & 0.85 & 0.96 & 0.29 & \multirow{2}{*}{9.5} \\
{} & {} & {(8)} & {(10)} & {(9)} & {(10)} & {(9)} & {(8)} & {(10)} & \\ 
\hline\hline\hline
% TEACHER REFERENCE
% {"BERTScore/precision": 1.0000000323131408, "BERTScore/recall": 1.0000000323131408, "BERTScore/f1": 1.0000000323131408, "BERTScore/hashcode": "distilbert-base-uncased_L5_no-idf_version=0.3.12(hug_trans=4.28.1)", "DialogRPT/_score/max": 0.6884973049163818, "DialogRPT/_score/avg": 0.3187272720339097, "DialogRPT/human_vs_rand/max": 0.9999984502792358, "DialogRPT/human_vs_rand/avg": 0.8635582583666448, "DialogRPT/human_vs_machine/max": 0.9999935626983643, "DialogRPT/human_vs_machine/avg": 0.9914906756345169, "DialogRPT/updown/max": 0.6670029163360596, "DialogRPT/updown/avg": 0.36568315894830794, "DialogRPT/depth/max": 0.9587414264678955, "DialogRPT/depth/avg": 0.7618460722895333, "DialogRPT/width/max": 0.9591518640518188, "DialogRPT/width/avg": 0.7867106894652048, "DialogRPT/final/max": 0.6884973049163818, "DialogRPT/final/avg": 0.3187272720339097}
{\sc teacher} & {\sc reference} & 1.00 & 1.00 & 1.00 & 0.37 & 0.86 & 0.99 & 0.32 & {} \\
\hline
\end{tabularx}
\caption{Leaderboard for the evaluation phase with scores and ranks for BERTScore (P = precision, R = recall) and DialogRPT (U = updown, HvR = human vs. rand, HvM = human vs. machine).}
\label{tab:automated-metrics}
\end{table*}

As can be observed from \Cref{tab:automated-metrics}, the NAISTeacher system \citep{vasselli_naisteacher_2023} attained the highest average rank on BERTScore and DialogRPT. On average, the responses were the closest to the teacher's response, the most relevant for the given dialogue context, and also the most likely to be human-written. The system also achieved the second-best result on the DialogRPT updown metric, which indicated that the generated responses were likely to receive upvotes. In addition to achieving the best average rank on the evaluation metrics, the system also achieved the best rank on all three criteria of pedagogical ability evaluated by human raters (see \Cref{fig:human-evaluation}). In particular, the responses were found to be the most helpful overall.

\Cref{tab:automated-metrics} further shows that the best result on the DialogRPT updown metric was achieved by the Cornell team \citep{hicke_assessing_2023}. The responses generated by GPT-4 were the most likely to receive upvotes on average ($0.52$) when they were submitted with a teacher prefix. However, when the team submitted the same responses \emph{without} the prefix, they received a much lower score ($0.4$) and ranked sixth on the same metric. This remarkable outcome highlighted the unanticipated sensitivity of the DialogRPT metric to the presence or absence of a prefix. 

The ADAIO system \citep{adigwe_promptbased_2023} attained the second-best average rank in both the automated evaluation phase (\Cref{tab:automated-metrics}) and the human evaluation phase (\Cref{fig:human-evaluation}). The results indicated that the use of well-engineered prompts, including good teaching examples (NAISTeacher, \#1) or teaching approaches and goals (ADAIO, \#2), resulted in a high rank on BERTScore, DialogRPT, and assessment of pedagogic ability.

It is interesting to note that the teacher's response was ranked \emph{lower} than the top three systems built on GPT-3 and GPT-4 (\Cref{fig:human-evaluation}), which contradicts the results of \citet{tack_ai_2022}. This striking observation might be explained by some differences in the human evaluation procedure: while any native English speaker could participate in \citet{tack_ai_2022}, only raters working in education and training could participate in the shared task. Some of these raters gave specific feedback stating that they found the non-standard language used by the teacher in the chatroom (including typos, spelling mistakes, lowercasing, etc.) less professional.

For more in-depth analyses, the reader is referred to the system papers cited in this paper. In these articles, the participating teams ran additional analyses and made critical observations. For example, \citeauthor{baladon_retuytinco_2023} (RETUYT-InCo) observed that fine-tuned models obtained better results on BERTScore, prompting obtained better results on DialogRPT, and methods that combined both techniques showed competitive results across all metrics. At the same time, they found that a baseline generating ``Hello'' in response to every prompt achieved the best result for BERTScore precision and DialogRPT updown. 
\citeauthor{huber_enhancing_2023} (Data Science-NLP-HSG) found that GPT-2—a smaller model with $124$ million parameters—achieved competitive performance compared to the T5-base model. Moreover, they found that, even though they maximized BERTScore F1 as a reward function, their model scored highly in the other evaluation metrics.
\citeauthor{vasselli_naisteacher_2023} (NAIST) noted that DialogRPT often preferred complete answers that were not very teacher-like over responses that helped the student find the answer by themselves.

\section{Discussion}

Although the inaugural shared task on generating AI teacher responses in educational dialogues can be considered a success, the results demonstrate that the evaluation of natural language generation models remains challenging. Ultimately, we would like to have at our disposal precise, valid, and—ideally—automated methods that reward machines and/or humans for their pedagogical abilities. However, we are probably still a long way from achieving this ultimate goal.

The existing automated metrics are not capable of rewarding models for their ability to showcase pedagogical skills. In particular, to the best of our knowledge, there is no comprehensive metric capable of evaluating whether responses are likely to be produced by a teacher or whether they demonstrate an understanding of what the student is saying and are helping the student. Moreover, popular automated metrics such as BERTScore and DialogRPT used in this task show considerable sensitivity to construct-irrelevant variations, as demonstrated by the use of a ``Hello'' baseline~\citep{baladon_retuytinco_2023} and the inclusion of the ``teacher:'' prefix~\cite{hicke_assessing_2023}. Future editions of this task should, therefore, aim to develop or resort to more accurate and domain-specific automated metrics, following observations and suggestions from several competing teams \citep{adigwe_promptbased_2023,baladon_retuytinco_2023,hicke_assessing_2023,vasselli_naisteacher_2023}.

Due to the lack of adequate metrics, we need to resort to manual evaluation methods in order to achieve more precise assessments. However, a typical drawback to manual evaluation is that it is very costly and time-consuming to have a sufficient number of raters evaluating {\em any} possible response that can be generated in the large space of possible teacher replies. Due to practical and budgetary limitations, it is challenging to organize a shared task during which any possible number of submissions can, in principle, be evaluated with adequately remunerated human evaluations.

What is more, data is very important in the context of real-world applications and shared tasks. 
Although the corpus used in this shared task is a valuable resource in our domain, some particularities of this corpus and the data sampling method also undeniably impacted the results. Therefore, in future editions of this shared task, we should rethink some of the current potential limitations, such as the fact that the dialogues had to be limited to $100$ tokens, resulting in partial conversations; the fact that some dialogues, if randomly extracted from the data, might have led to data leakage; and the fact that the dialogues did not always follow a strictly role-alternating format, with some teacher turns preceded by previous teacher utterances rather than a student utterance.
%In the context of educational dialogues, TSCC corpus is one of the widely used ones, 
% However, as has been pointed out in Section~\ref{sec:data} and noted by several teams \citep{hicke_assessing_2023,baladon_retuytinco_2023,vasselli_naisteacher_2023}, the data format and characteristics of the corpus posed particular challenges for the task: the fact that the dialogues had to be limited to $100$ tokens, resulting in some conversations being incomplete; the fact that some dialogues, if extracted from the data randomly might have led to data leakage; and the fact that the dialogues did not always follow strictly role-alternating format, with some teacher turns being preceded by previous teacher utterance, rather than a student utterance.

In summary, the field of education has already been significantly changed by LLMs, whose capabilities continue to improve constantly. We hope that this shared task will help the scientific community better understand the current capabilities of LLMs in educational contexts. Having learned from this shared task and going forward, we hope to make its future iterations even more informative.

\section{Conclusion}

%TODO: Summarize what we've learned from this round and what could be improved next time. 

The primary goal of this shared task was to explore the potential of the current state-of-the-art NLP and AI methods in generating teacher responses in the context of real-world teacher-student interactions. Several strong and diverse teams participated in the task and submitted outputs of their systems to the competition, and even more researchers expressed their interest. The teams used a variety of state-of-the-art large language models and explored diverse prompting and fine-tuning approaches. 
Importantly, these results not only shed light on the current state-of-the-art on this task, but also highlighted some critical limitations that should be addressed in the future.

\section*{Acknowledgements}

We thank the participants for their submissions and active involvement in this shared task. We are also grateful to them for the detailed and helpful peer reviews they provided to other shared task participants. Finally, we thank the anonymous raters on Prolific for taking the time to provide us with additional feedback.

% Entries for the entire Anthology, followed by custom entries
\bibliography{references,system-papers}

\begin{thebibliography}{29}
\expandafter\ifx\csname natexlab\endcsname\relax\def\natexlab#1{#1}\fi

\bibitem[{Adigwe and Yuan(2023)}]{adigwe_promptbased_2023}
Adaeze Adigwe and Zheng Yuan. 2023.
\newblock The {ADAIO} system at the {BEA-2023 Shared Task}: Shared task
  generating {AI} teacher responses in educational dialogues.
\newblock In \emph{Proceedings of the 18th {{Workshop}} on {{Innovative Use}}
  of {{NLP}} for {{Building Educational Applications}}}, page to appear,
  {Toronto, Canada}. {Association for Computational Linguistics}.

\bibitem[{Balad{\'o}n et~al.(2023)Balad{\'o}n, Sastre, Chiruzzo, and
  Ros{\'a}}]{baladon_retuytinco_2023}
Alexis Balad{\'o}n, Ignacio Sastre, Luis Chiruzzo, and Aiala Ros{\'a}. 2023.
\newblock {RETUYT-InCo} at {BEA 2023 Shared Task}: Tuning open-source {LLMs}
  for generating teacher responses.
\newblock In \emph{Proceedings of the 18th {{Workshop}} on {{Innovative Use}}
  of {{NLP}} for {{Building Educational Applications}}}, page to appear,
  {Toronto, Canada}. {Association for Computational Linguistics}.

\bibitem[{Bibauw et~al.(2022)Bibauw, {Van den Noortgate}, François, and
  Desmet}]{bibauw_dialogue_2022}
Serge Bibauw, Wim {Van den Noortgate}, Thomas François, and Piet Desmet. 2022.
\newblock \href {https://doi.org/10125/73488} {Dialogue systems for language
  learning: A meta-analysis}.
\newblock \emph{Language Learning \& Technology}, 26(1):1--24.

\bibitem[{Bommasani et~al.(2021)Bommasani, Hudson, Adeli, Altman, Arora, {von
  Arx}, Bernstein, Bohg, Bosselut, Brunskill, Brynjolfsson, Buch, Card,
  Castellon, Chatterji, Chen, Creel, Davis, Demszky, Donahue, Doumbouya,
  Durmus, Ermon, Etchemendy, Ethayarajh, {Fei-Fei}, Finn, Gale, Gillespie,
  Goel, Goodman, Grossman, Guha, Hashimoto, Henderson, Hewitt, Ho, Hong, Hsu,
  Huang, Icard, Jain, Jurafsky, Kalluri, Karamcheti, Keeling, Khani, Khattab,
  Kohd, Krass, Krishna, Kuditipudi, Kumar, Ladhak, Lee, Lee, Leskovec, Levent,
  Li, Li, Ma, Malik, Manning, Mirchandani, Mitchell, Munyikwa, Nair, Narayan,
  Narayanan, Newman, Nie, Niebles, Nilforoshan, Nyarko, Ogut, Orr,
  Papadimitriou, Park, Piech, Portelance, Potts, Raghunathan, Reich, Ren, Rong,
  Roohani, Ruiz, Ryan, R{\'e}, Sadigh, Sagawa, Santhanam, Shih, Srinivasan,
  Tamkin, Taori, Thomas, Tram{\`e}r, Wang, Wang, Wu, Wu, Wu, Xie, Yasunaga,
  You, Zaharia, Zhang, Zhang, Zhang, Zhang, Zheng, Zhou, and
  Liang}]{bommasani_opportunities_2021}
Rishi Bommasani, Drew~A. Hudson, Ehsan Adeli, Russ Altman, Simran Arora, Sydney
  {von Arx}, Michael~S. Bernstein, Jeannette Bohg, Antoine Bosselut, Emma
  Brunskill, Erik Brynjolfsson, Shyamal Buch, Dallas Card, Rodrigo Castellon,
  Niladri Chatterji, Annie Chen, Kathleen Creel, Jared~Quincy Davis, Dora
  Demszky, Chris Donahue, Moussa Doumbouya, Esin Durmus, Stefano Ermon, John
  Etchemendy, Kawin Ethayarajh, Li~{Fei-Fei}, Chelsea Finn, Trevor Gale, Lauren
  Gillespie, Karan Goel, Noah Goodman, Shelby Grossman, Neel Guha, Tatsunori
  Hashimoto, Peter Henderson, John Hewitt, Daniel~E. Ho, Jenny Hong, Kyle Hsu,
  Jing Huang, Thomas Icard, Saahil Jain, Dan Jurafsky, Pratyusha Kalluri,
  Siddharth Karamcheti, Geoff Keeling, Fereshte Khani, Omar Khattab, Pang~Wei
  Kohd, Mark Krass, Ranjay Krishna, Rohith Kuditipudi, Ananya Kumar, Faisal
  Ladhak, Mina Lee, Tony Lee, Jure Leskovec, Isabelle Levent, Xiang~Lisa Li,
  Xuechen Li, Tengyu Ma, Ali Malik, Christopher~D. Manning, Suvir Mirchandani,
  Eric Mitchell, Zanele Munyikwa, Suraj Nair, Avanika Narayan, Deepak
  Narayanan, Ben Newman, Allen Nie, Juan~Carlos Niebles, Hamed Nilforoshan,
  Julian Nyarko, Giray Ogut, Laurel Orr, Isabel Papadimitriou, Joon~Sung Park,
  Chris Piech, Eva Portelance, Christopher Potts, Aditi Raghunathan, Rob Reich,
  Hongyu Ren, Frieda Rong, Yusuf Roohani, Camilo Ruiz, Jack Ryan, Christopher
  R{\'e}, Dorsa Sadigh, Shiori Sagawa, Keshav Santhanam, Andy Shih, Krishnan
  Srinivasan, Alex Tamkin, Rohan Taori, Armin~W. Thomas, Florian Tram{\`e}r,
  Rose~E. Wang, William Wang, Bohan Wu, Jiajun Wu, Yuhuai Wu, Sang~Michael Xie,
  Michihiro Yasunaga, Jiaxuan You, Matei Zaharia, Michael Zhang, Tianyi Zhang,
  Xikun Zhang, Yuhui Zhang, Lucia Zheng, Kaitlyn Zhou, and Percy Liang. 2021.
\newblock \href {http://arxiv.org/abs/2108.07258} {On the opportunities and
  risks of foundation models}.
\newblock Technical report, {Stanford University}, {Center for Research on
  Foundation Models (CRFM)}.

\bibitem[{Brown et~al.(2020)Brown, Mann, Ryder, Subbiah, Kaplan, Dhariwal,
  Neelakantan, Shyam, Sastry, Askell, Agarwal, Herbert-Voss, Krueger, Henighan,
  Child, Ramesh, Ziegler, Wu, Winter, Hesse, Chen, Sigler, Litwin, Gray, Chess,
  Clark, Berner, McCandlish, Radford, Sutskever, and
  Amodei}]{brown2020language}
Tom Brown, Benjamin Mann, Nick Ryder, Melanie Subbiah, Jared~D Kaplan, Prafulla
  Dhariwal, Arvind Neelakantan, Pranav Shyam, Girish Sastry, Amanda Askell,
  Sandhini Agarwal, Ariel Herbert-Voss, Gretchen Krueger, Tom Henighan, Rewon
  Child, Aditya Ramesh, Daniel Ziegler, Jeffrey Wu, Clemens Winter, Chris
  Hesse, Mark Chen, Eric Sigler, Mateusz Litwin, Scott Gray, Benjamin Chess,
  Jack Clark, Christopher Berner, Sam McCandlish, Alec Radford, Ilya Sutskever,
  and Dario Amodei. 2020.
\newblock \href
  {https://proceedings.neurips.cc/paper_files/paper/2020/file/1457c0d6bfcb4967418bfb8ac142f64a-Paper.pdf}
  {{Language models are few-shot learners}}.
\newblock In \emph{Advances in Neural Information Processing Systems},
  volume~33, pages 1877--1901. {Curran Associates}.

\bibitem[{Caines et~al.(2022)Caines, Yannakoudakis, Allen, Pérez-Paredes,
  Byrne, and Buttery}]{caines_teacherstudent_2022}
Andrew Caines, Helen Yannakoudakis, Helen Allen, Pascual Pérez-Paredes, Bill
  Byrne, and Paula Buttery. 2022.
\newblock \href {https://aclanthology.org/2022.nlp4call-1.3} {The
  {{Teacher-Student Chatroom Corpus}} version 2: More lessons, new annotation,
  automatic detection of sequence shifts}.
\newblock In \emph{Proceedings of the 11th {{Workshop}} on {{NLP}} for
  {{Computer Assisted Language Learning}}}, pages 23--35, {Louvain-la-Neuve,
  Belgium}. {LiU Electronic Press}.

\bibitem[{Caines et~al.(2020)Caines, Yannakoudakis, Edmondson, Allen,
  Pérez-Paredes, Byrne, and Buttery}]{caines_teacherstudent_2020}
Andrew Caines, Helen Yannakoudakis, Helena Edmondson, Helen Allen, Pascual
  Pérez-Paredes, Bill Byrne, and Paula Buttery. 2020.
\newblock \href {https://aclanthology.org/2020.nlp4call-1.2} {The
  teacher-student chatroom corpus}.
\newblock In \emph{Proceedings of the 9th Workshop on {{NLP}} for Computer
  Assisted Language Learning}, pages 10--20, {Gothenburg, Sweden}. {LiU
  Electronic Press}.

\bibitem[{Chung et~al.(2022)Chung, Hou, Longpre, Zoph, Tay, Fedus, Li, Wang,
  Dehghani, Brahma, Webson, Gu, Dai, Suzgun, Chen, Chowdhery, Castro-Ros,
  Pellat, Robinson, Valter, Narang, Mishra, Yu, Zhao, Huang, Dai, Yu, Petrov,
  Chi, Dean, Devlin, Roberts, Zhou, Le, and Wei}]{chung2022scaling}
Hyung~Won Chung, Le~Hou, Shayne Longpre, Barret Zoph, Yi~Tay, William Fedus,
  Yunxuan Li, Xuezhi Wang, Mostafa Dehghani, Siddhartha Brahma, Albert Webson,
  Shixiang~Shane Gu, Zhuyun Dai, Mirac Suzgun, Xinyun Chen, Aakanksha
  Chowdhery, Alex Castro-Ros, Marie Pellat, Kevin Robinson, Dasha Valter,
  Sharan Narang, Gaurav Mishra, Adams Yu, Vincent Zhao, Yanping Huang, Andrew
  Dai, Hongkun Yu, Slav Petrov, Ed~H. Chi, Jeff Dean, Jacob Devlin, Adam
  Roberts, Denny Zhou, Quoc~V. Le, and Jason Wei. 2022.
\newblock \href {https://doi.org/10.48550/arXiv.2210.11416} {Scaling
  instruction-finetuned language models}.
\newblock arXiv:2210.11416.

\bibitem[{Gao et~al.(2020)Gao, Zhang, Galley, Brockett, and
  Dolan}]{gao_dialogue_2020}
Xiang Gao, Yizhe Zhang, Michel Galley, Chris Brockett, and Bill Dolan. 2020.
\newblock \href {https://doi.org/10.18653/v1/2020.emnlp-main.28} {Dialogue
  response ranking training with large-scale human feedback data}.
\newblock In \emph{Proceedings of the 2020 {{Conference}} on {{Empirical
  Methods}} in {{Natural Language Processing}} ({{EMNLP}})}, pages 386--395,
  {Online}. {Association for Computational Linguistics}.

\bibitem[{Hicke et~al.(2023)Hicke, Masand, Guo, and
  Gangavarapu}]{hicke_assessing_2023}
Yann Hicke, Abhishek Masand, Wentao Guo, and Tushaar Gangavarapu. 2023.
\newblock Assessing the efficacy of large language models in generating
  accurate teacher responses.
\newblock In \emph{Proceedings of the 18th {{Workshop}} on {{Innovative Use}}
  of {{NLP}} for {{Building Educational Applications}}}, page to appear,
  {Toronto, Canada}. {Association for Computational Linguistics}.

\bibitem[{Hu et~al.(2021)Hu, Shen, Wallis, Allen-Zhu, Li, Wang, Wang, and
  Chen}]{hu2021lora}
Edward~J. Hu, Yelong Shen, Phillip Wallis, Zeyuan Allen-Zhu, Yuanzhi Li, Shean
  Wang, Lu~Wang, and Weizhu Chen. 2021.
\newblock \href {https://doi.org/10.48550/arXiv.2106.09685} {{LoRA}: Low-rank
  adaptation of large language models}.
\newblock arXiv:2106.09685.

\bibitem[{Huber et~al.(2023)Huber, Niklaus, and
  Handschuh}]{huber_enhancing_2023}
Thomas Huber, Christina Niklaus, and Siegfried Handschuh. 2023.
\newblock Enhancing educational dialogues: A reinforcement learning approach
  for generating {AI} teacher responses.
\newblock In \emph{Proceedings of the 18th {{Workshop}} on {{Innovative Use}}
  of {{NLP}} for {{Building Educational Applications}}}, page to appear,
  {Toronto, Canada}. {Association for Computational Linguistics}.

\bibitem[{Kuhn(1955)}]{kuhn_hungarian_1955}
H.~W. Kuhn. 1955.
\newblock \href {https://doi.org/10.1002/nav.3800020109} {The {{Hungarian}}
  method for the assignment problem}.
\newblock \emph{Naval Research Logistics Quarterly}, 2(1-2):83--97.

\bibitem[{{Le Scao} et~al.(2022){Le Scao}, {Fan}, {Akiki}, {Pavlick}, {Ilić},
  {Hesslow}, {Castagné}, {Sasha Luccioni}, {Yvon}, {Gallé}
  et~al.}]{scao2022bloom}
Teven {Le Scao}, Angela {Fan}, Christopher {Akiki}, Ellie {Pavlick}, Suzana
  {Ilić}, Daniel {Hesslow}, Roman {Castagné}, Alexandra {Sasha Luccioni},
  François {Yvon}, Matthias {Gallé}, et~al. 2022.
\newblock \href {https://doi.org/10.48550/arXiv.2211.05100} {{BLOOM}: A
  {176B}-parameter open-access multilingual language model}.
\newblock arXiv:2211.05100.

\bibitem[{Li et~al.(2019)Li, Weston, and Roller}]{li_acuteeval_2019}
Margaret Li, Jason Weston, and Stephen Roller. 2019.
\newblock \href {https://doi.org/10.48550/arXiv.1909.03087} {{ACUTE-EVAL}:
  Improved dialogue evaluation with optimized questions and multi-turn
  comparisons}.
\newblock arXiv:1909.03087.

\bibitem[{Omidvar and An(2023)}]{omidvar_empowering_2023}
Amin Omidvar and Aijun An. 2023.
\newblock Empowering conversational agents using semantic in-context learning.
\newblock In \emph{Proceedings of the 18th {{Workshop}} on {{Innovative Use}}
  of {{NLP}} for {{Building Educational Applications}}}, page to appear,
  {Toronto, Canada}. {Association for Computational Linguistics}.

\bibitem[{Pavao et~al.(2022)Pavao, Guyon, Letournel, Bar{\'o}, Escalante,
  Escalera, Thomas, and Xu}]{codalab_competitions}
Adrien Pavao, Isabelle Guyon, Anne-Catherine Letournel, Xavier Bar{\'o}, Hugo
  Escalante, Sergio Escalera, Tyler Thomas, and Zhen Xu. 2022.
\newblock {{CodaLab Competitions}}: {{An}} open source platform to organize
  scientific challenges.
\newblock \emph{Technical report}.

\bibitem[{Pillutla et~al.(2021)Pillutla, Swayamdipta, Zellers, Thickstun,
  Welleck, Choi, and Harchaoui}]{pillutla_mauve_2021}
Krishna Pillutla, Swabha Swayamdipta, Rowan Zellers, John Thickstun, Sean
  Welleck, Yejin Choi, and Zaid Harchaoui. 2021.
\newblock \href {http://arxiv.org/abs/2102.01454} {{MAUVE}: Measuring the gap
  between neural text and human text using divergence frontiers}.
\newblock In \emph{Advances in {{Neural Information Processing Systems}} 34
  Pre-Proceedings ({{NeurIPS}} 2021)}, pages 1--35.

\bibitem[{Radford et~al.(2019)Radford, Wu, Child, Luan, Amodei, and
  Sutskever}]{radford2019language}
Alec Radford, Jeffrey Wu, Rewon Child, David Luan, Dario Amodei, and Ilya
  Sutskever. 2019.
\newblock \href
  {https://d4mucfpksywv.cloudfront.net/better-language-models/language-models.pdf}
  {{Language models are unsupervised multitask learners}}.
\newblock OpenAI blog.

\bibitem[{Raffel et~al.(2020)Raffel, Shazeer, Roberts, Lee, Narang, Matena,
  Zhou, Li, and Liu}]{raffel2020exploring}
Colin Raffel, Noam Shazeer, Adam Roberts, Katherine Lee, Sharan Narang, Michael
  Matena, Yanqi Zhou, Wei Li, and Peter~J Liu. 2020.
\newblock \href {http://jmlr.org/papers/v21/20-074.html} {{Exploring the limits
  of transfer learning with a unified text-to-text transformer}}.
\newblock \emph{Journal of Machine Learning Research}, 21(140):1--67.

\bibitem[{Ramamurthy et~al.(2023)Ramamurthy, Ammanabrolu, Brantley, Hessel,
  Sifa, Bauckhage, Hajishirzi, and Choi}]{ramamurthy2022reinforcement}
Rajkumar Ramamurthy, Prithviraj Ammanabrolu, Kianté Brantley, Jack Hessel,
  Rafet Sifa, Christian Bauckhage, Hannaneh Hajishirzi, and Yejin Choi. 2023.
\newblock \href {http://arxiv.org/abs/2210.01241} {Is reinforcement learning
  (not) for natural language processing: Benchmarks, baselines, and building
  blocks for natural language policy optimization}.
\newblock arXiv:2210.01241.

\bibitem[{Tack and Piech(2022)}]{tack_ai_2022}
Ana{\"i}s Tack and Chris Piech. 2022.
\newblock \href {https://doi.org/10.5281/zenodo.6853187} {The {{AI Teacher
  Test}}: {{Measuring}} the pedagogical ability of {{Blender}} and {{GPT-3}} in
  educational dialogues}.
\newblock In \emph{Proceedings of the 15th {{International Conference}} on
  {{Educational Data Mining}}}, volume~15, pages 522--529, {Durham, United
  Kingdom}. {International Educational Data Mining Society}.

\bibitem[{Taori et~al.(2023)Taori, Gulrajani, Zhang, Dubois, Li, Guestrin,
  Liang, and Hashimoto}]{taori2023stanford}
Rohan Taori, Ishaan Gulrajani, Tianyi Zhang, Yann Dubois, Xuechen Li, Carlos
  Guestrin, Percy Liang, and Tatsunori~B. Hashimoto. 2023.
\newblock \href {https://github.com/tatsu-lab/stanford_alpaca} {{Stanford
  Alpaca}: An instruction-following {LLaMA} model}.
\newblock GitHub.

\bibitem[{Touvron et~al.(2023)Touvron, Lavril, Izacard, Martinet, Lachaux,
  Lacroix, Rozière, Goyal, Hambro, Azhar, Rodriguez, Joulin, Grave, and
  Lample}]{touvron2023llama}
Hugo Touvron, Thibaut Lavril, Gautier Izacard, Xavier Martinet, Marie-Anne
  Lachaux, Timothée Lacroix, Baptiste Rozière, Naman Goyal, Eric Hambro,
  Faisal Azhar, Aurelien Rodriguez, Armand Joulin, Edouard Grave, and Guillaume
  Lample. 2023.
\newblock \href {http://arxiv.org/abs/2302.13971} {{LLaMA}: Open and efficient
  foundation language models}.
\newblock arXiv:2302.13971.

\bibitem[{Vasselli et~al.(2023)Vasselli, Vasselli, Nohejl, and
  Watanabe}]{vasselli_naisteacher_2023}
Justin Vasselli, Christopher Vasselli, Adam Nohejl, and Taro Watanabe. 2023.
\newblock {NAISTeacher}: A prompt and rerank approach to generating teacher
  utterances in educational dialogues.
\newblock In \emph{Proceedings of the 18th {{Workshop}} on {{Innovative Use}}
  of {{NLP}} for {{Building Educational Applications}}}, page to appear,
  {Toronto, Canada}. {Association for Computational Linguistics}.

\bibitem[{Wollny et~al.(2021)Wollny, Schneider, Di~Mitri, Weidlich, Rittberger,
  and Drachsler}]{wollny_are_2021}
Sebastian Wollny, Jan Schneider, Daniele Di~Mitri, Joshua Weidlich, Marc
  Rittberger, and Hendrik Drachsler. 2021.
\newblock \href {https://doi.org/10.3389/frai.2021.654924} {Are we there yet? -
  a systematic literature review on chatbots in education}.
\newblock \emph{Frontiers in Artificial Intelligence}, 4:654924.

\bibitem[{Yeh et~al.(2021)Yeh, Eskenazi, and Mehri}]{yeh_comprehensive_2021}
Yi-Ting Yeh, Maxine Eskenazi, and Shikib Mehri. 2021.
\newblock \href {https://doi.org/10.18653/v1/2021.eancs-1.3} {A {{Comprehensive
  Assessment}} of {{Dialog Evaluation Metrics}}}.
\newblock In \emph{The {{First Workshop}} on {{Evaluations}} and
  {{Assessments}} of {{Neural Conversation Systems}}}, pages 15--33, {Online}.
  {Association for Computational Linguistics}.

\bibitem[{Zhang et~al.(2020)Zhang, Kishore, Wu, Weinberger, and
  Artzi}]{zhang*_bertscore_2020}
Tianyi Zhang, Varsha Kishore, Felix Wu, Kilian~Q. Weinberger, and Yoav Artzi.
  2020.
\newblock \href {https://iclr.cc/virtual_2020/poster_SkeHuCVFDr.html}
  {{{BERTScore}}: {{Evaluating}} text generation with {{BERT}}}.
\newblock In \emph{International {{Conference}} on {{Learning
  Representations}}}, Online.

\bibitem[{Zhang et~al.()Zhang, Sun, Galley, Chen, Brockett, Gao, Gao, Liu, and
  Dolan}]{zhang2019dialogpt}
Yizhe Zhang, Siqi Sun, Michel Galley, Yen-Chun Chen, Chris Brockett, Xiang Gao,
  Jianfeng Gao, Jingjing Liu, and Bill Dolan.
\newblock \href {https://doi.org/10.18653/v1/2020.acl-demos.30} {{DIALOGPT}:
  Large-scale generative pre-training for conversational response generation}.
\newblock In \emph{Proceedings of the 58th {{Annual Meeting}} of the
  {{Association}} for {{Computational Linguistics}}: {{System
  Demonstrations}}}, pages 270--278. {Association for Computational
  Linguistics}.

\end{thebibliography}

\end{document}